# Fine-Grained Japanese Clinical Named Entity Recognition: A Controlled Comparison of BERT and Open-Weight Large Language Models

Feng Han[1]*, Ayako Yagahara[2,3]*, Kenji Hirata[4], Atsushi Yoshida[1,5], and Kohsuke Kudo[5]

[1] Division of AI Support for Medical Research, Research Center for Cooperative Projects, Faculty of Medicine, Hokkaido University, Sapporo, Japan

[2] Faculty of Health Sciences, Hokkaido University of Science, Sapporo, Japan

[3] Faculty of Health Sciences, Hokkaido University, Sapporo, Japan

[4] Department of Diagnostic Imaging, Faculty of Medicine, Hokkaido University, Sapporo, Japan; Present address: Department of Nuclear Medicine, Faculty of Medicine, Kanazawa University, Kanazawa, Japan

[5] Department of Diagnostic Imaging, Faculty of Medicine, Hokkaido University, Sapporo, Japan

**Corresponding Authors**

*Ayako Yagahara and Feng Han are co-corresponding authors.*

**Correspondence to:**

Ayako Yagahara — Email: yagahara-a@hus.ac.jp

Feng Han — Email: hanfenghokudai@gmail.com

## Abstract

Background: Generative large language models (LLMs) offer an alternative to encoder-based clinical named entity recognition (NER), but their performance on fine-grained Japanese extraction tasks and sensitivity to generation settings remain uncertain.

Methods: We compared two fine-tuned BERT–CRF baselines with nine open-weight LLMs on iCorpus, which defines 113 entity categories. The evaluation used 433 sentences from 36 test documents; 1,672 sentences were used for training. LLMs were evaluated using zero-shot prompting with and without JSON Schema constraints and three-seed QLoRA fine-tuning. Fine-tuned models were assessed under different output-token limits, including 900 and larger model-specific limits of 2,600 or 4,096 tokens. Performance was measured using sentence-level label–surface matching with relaxed and exact-surface F1.

Results: The highest zero-shot relaxed F1 was 0.337, compared with 0.836–0.841 for the BERT baselines. At the larger output limits, all eight fine-tuned LLMs with at least 13B total parameters exceeded both BERT baselines (relaxed F1, 0.882–0.914; exact-surface F1, 0.840–0.880). Across three-seed means, increasing the output limit improved relaxed F1 by 0.052–0.085, alongside higher recall and fewer length-limited outputs. JSON Schema constraints changed relaxed F1 by −0.014 to +0.008 among seven models outside the two reasoning-oriented configurations, while medical–general model pairs showed no consistent advantage for medical specialization.

Conclusions: On this historical Japanese case-report benchmark, task-specific fine-tuning combined with adequate generation capacity was associated with stronger LLM extraction performance. Output-token limits materially affected comparisons with BERT, whereas enforcing output structure alone did not overcome low zero-shot accuracy.

*Keywords: clinical natural language processing; named entity recognition; Japanese; large language models; BERT; structured output; generation budget*

# 1. Introduction

Named entity recognition (NER) is a foundational step in converting clinical narratives into structured information. Clinical entity extraction extends beyond identifying disease names to recognizing findings, examinations, treatments, medications, and other clinically relevant expressions. Fine-grained annotation schemes may further distinguish whether a finding concerns the patient or another person and whether it is affirmed, negated, or suspected. Such fine-grained distinctions require NER systems to identify diverse clinical expressions while accurately assigning them to clinically meaningful categories.

Pretrained encoders such as BERT have provided an established approach to supervised NER through task-specific sequence labeling (Devlin et al., 2019). For Japanese clinical text, UTH-BERT offers an encoder pretrained on clinical records, providing domain-specific representations for downstream applications (Kawazoe et al., 2021). More recently, large language models (LLMs) have introduced an alternative approach by formulating entity extraction as instruction-guided text generation rather than token-level classification (Wang et al., 2025). The growing availability of open-weight Japanese-language and medically specialized LLMs has further expanded the range of models available for clinical NER research (Sellergren et al., 2025; Aizawa et al., 2026). However, the relative extraction performance of generative LLMs and established encoder-based methods remains an important question, particularly for fine-grained clinical NER.

Previous studies have reported varying results for generative LLMs and supervised encoder-based models in clinical and biomedical NER. In multilingual clinical NER, prompt-based generative models have underperformed supervised encoders (Naguib et al., 2024), whereas instruction tuning has enabled general-purpose LLMs to achieve performance comparable to specialized biomedical encoders (Keloth et al., 2024).

Outside Japanese clinical benchmarks, task-oriented generative NER systems have also demonstrated the value of specialized adaptation: UniversalNER used targeted distillation for open NER (Zhou et al., 2024), whereas GNER incorporated negative training instances to improve generalization to unseen entity domains (Ding et al., 2024). In Japanese, Shimizu et al. (2025) reported that a fine-tuned LLaMA-3.1 model outperformed a BERT baseline for disease name recognition. These findings highlight the importance of task-specific adaptation, but their applicability to fine-grained Japanese clinical NER remains unclear.

Beyond task adaptation, generative NER introduces additional challenges related to output generation. Unlike sequence-labeling systems, which assign labels to spans of input text, generative LLMs must explicitly produce the extracted entities in their responses. Previous research using English clinical texts has shown that the requested output format can influence extraction performance (del Moral-González et al., 2025), while grammar-constrained decoding has been investigated as a means of enforcing structured outputs in information extraction tasks (Geng et al., 2023). However, a correctly formatted response does not necessarily contain all relevant entities. When clinical text contains numerous entities, the required output can become lengthy, and responses may be truncated by the generation token limit, resulting in incomplete extraction. This dependence on output length also complicates comparisons with encoder-based systems, which do not need to generate an explicit list of extracted entities. The effects of output-format constraints and generation budgets therefore warrant further investigation in fine-grained Japanese clinical NER.

The Japanese case-report corpus iCorpus provides a setting for addressing these questions. Its annotation scheme comprises 113 entity categories, covering a wide range of clinical concepts and distinctions related to experiencer, polarity, and change (Kawazoe and Shinohara, 2022; Shibata et al., 2024). This comprehensive annotation scheme enables evaluation beyond disease name recognition, while the presence of numerous entity mentions in individual sentences also makes output completeness an important consideration for generative extraction. The benchmark also provides an opportunity to compare medically specialized and general-purpose LLMs, although such comparisons cannot isolate the effects of domain-specific pretraining.

Although prior research established supervised entity and relation extraction on iCorpus (Shibata et al., 2024) and compared fine-tuned LLMs with BERT for Japanese disease-name recognition (Shimizu et al., 2025), their findings do not resolve how open-weight generative models compare with encoder-based systems on a shared, broadly annotated Japanese clinical NER benchmark. Moreover, output-format enforcement and generation-token limits can affect generative predictions independently of model weights, potentially influencing the interpretation of model comparisons. A systematic evaluation that explicitly examines task adaptation, decoding constraints, and output capacity is therefore needed to interpret BERT–LLM performance differences.

In this study, we compare two Japanese BERT-based systems with nine open-weight LLMs on iCorpus. Our primary objective is to evaluate the performance of LLMs relative to supervised BERT baselines in fine-grained Japanese clinical NER under zero-shot and task-specific fine-tuning settings (RQ1). We further examine the effects of JSON Schema-constrained decoding on zero-shot extraction (RQ2), the influence of generation budgets on the accuracy and completeness of fine-tuned LLM outputs (RQ3), and performance differences across model configurations, including comparisons between medically specialized and general-purpose LLMs (RQ4).

# 2. Methods

This study compared BERT-based sequence labeling and generative LLM approaches for fine-grained Japanese clinical NER using iCorpus. We evaluated task-specific fine-tuning, zero-shot extraction, output-format constraints, generation budgets, and model configurations. Figure 1 summarizes the experimental workflow.

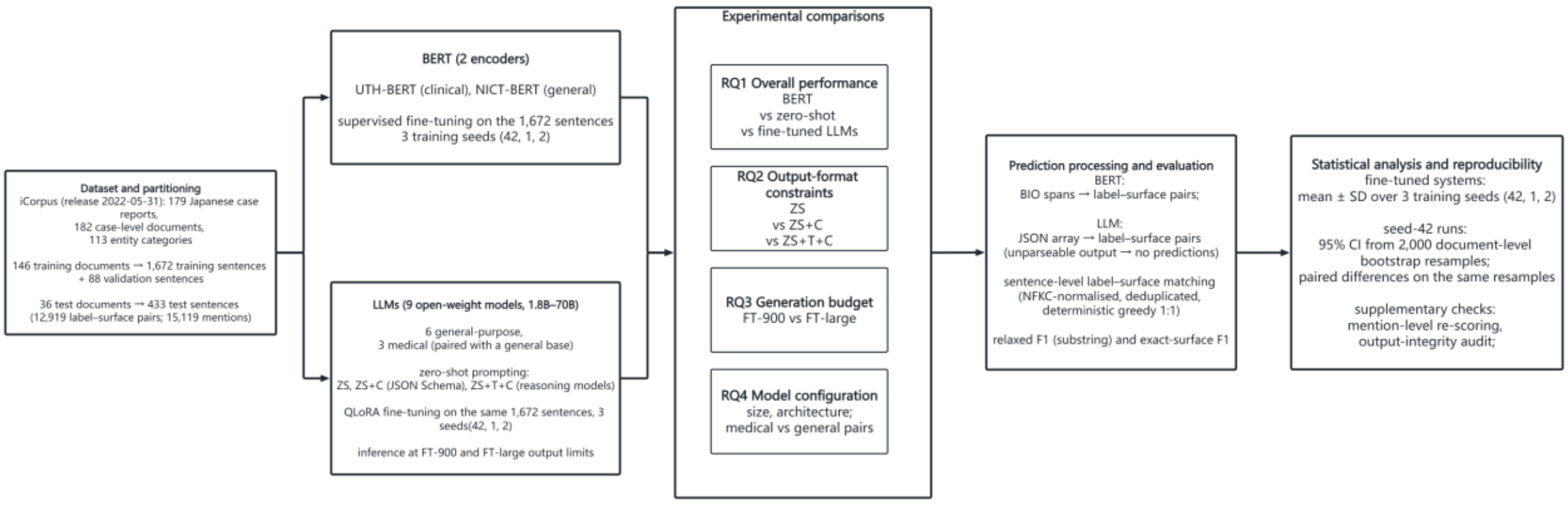

Figure 1. Study overview, from corpus partitioning and model training to the four experimental comparisons and common evaluation.

## 2.1. Dataset and Data Partitioning

We used iCorpus (released on May 31, 2022), a Japanese corpus of published case reports concerning designated intractable diseases (Kawazoe & Shinohara, 2022; Shibata et al., 2024; Department of Healthcare AI Development, 2022). The corpus contains character-level annotations covering 113 entity categories, including clinical concepts and attributes related to experiencer, polarity, and changes in patient condition.

The corpus consists of 179 case reports represented by 183 case-level files, as four reports contain two cases stored separately. In this study, each case-level file was treated as one document. One document without sentence segmentation was excluded, leaving 182 documents. We partitioned the dataset into 146 training documents (1,760 sentences) and 36 test documents (433 sentences). From the training partition, the final 88 sentences in corpus order were reserved for validation, resulting in 1,672 training sentences. The training and test partitions contained no overlapping documents.

## 2.2. NER Systems and Training

We evaluated two Japanese BERT-based models and nine open-weight LLMs covering general-purpose and medically specialized configurations, with LLM parameter counts from 1.8B to 70B and dense or mixture-of-experts (MoE) architectures (Table 1). Exact model identifiers are provided in Appendix F, Table F1. All supervised systems used the same designated training partition, and training was repeated with seeds 42, 1, and 2.

Table 1. Encoder-based and open-weight language models included in the main comparison.

| **Model** | **Domain** | **Architecture** | **Parameters** |
|---|---|---|---|
| UTH-BERT | Clinical | BERT–CRF | Base |
| NICT-BERT | General | BERT–CRF | Base |
| LLM-jp-3.1-1.8B | General | Dense | 1.8B |
| LLM-jp-3.1-13B | General | Dense | 13B |
| SIP-jmed-13B | Medical | Dense | 13B |
| Gemma-3-27B | General | Dense | 27B |
| MedGemma-27B | Medical | Dense | 27B |
| Qwen3-30B-A3B | General | MoE | 30B total; 3B active |
| LLM-jp-4-32B | General | MoE | 32B total; 3B active |
| Llama-3.3-70B | General | Dense | 70B |
| Med42-70B | Medical | Dense | 70B |

*For the MoE models, active parameter counts are approximate. Both BERT models and all nine LLMs were fine-tuned; only the LLMs also underwent zero-shot evaluation.* Model documentation: LLM-jp (2024), Gemma Team (2025), Qwen Team (2025), Grattafiori et al. (2024), and Christophe et al. (2024).

### 2.2.1. BERT-based NER

UTH-BERT, pretrained on Japanese clinical records (Kawazoe et al., 2021), and NICT-BERT, pretrained on Japanese Wikipedia (NICT, n.d.), were used as encoder baselines with a conditional random field (CRF) decoder. Input text was normalized using neologdn, and character-level entity annotations were converted to BIO tags aligned with model-specific subword tokenization. Sentences exceeding 510 subword tokens were divided at punctuation boundaries into windows, and predictions were concatenated to reconstruct sentence-level predictions, retaining all 433 test sentences.

BERT fine-tuning used AdamW (learning rate $2 \times 10^{-5}$; batch size 1) for a maximum of 250 epochs. Early stopping was based on validation micro-F1 with exact entity-boundary and type matching at subword positions, using patience of 15 epochs.

#### 2.2.2. LLM-based NER

LLM-based extraction was formulated as Japanese instruction-following generation. Each prompt contained the input sentence, extraction instructions, and a fixed inventory of 109 entity labels derived from the historical UTH-BERT annotation mapping (NER_tag2idx_0.json). This inventory was not identical to the 111 categories observed in the current training partition: three training categories (one reference entity each) were absent from the prompt list but remained in the fine-tuning reference responses, while one prompt category did not occur in the training partition. None of these four categories occurred in the test set. The prompt instructed models to copy minimal original entity expressions, omit absent entities, and return an empty array if none were found. Output was requested as JSON objects with label and text fields. The full prompt and label inventory are provided in Appendix A.

Fine-tuning examples paired the prompts with reference JSON arrays using each model's chat template. All nine LLMs were fine-tuned using QLoRA (Dettmers et al., 2023; Hu et al., 2022), with 4-bit NF4 quantization, LoRA rank 16, scaling factor 32, and dropout 0.05 applied to attention and MLP projections. AdamW was used with a learning rate of $2 \times 10^{-4}$, cosine scheduling, and an effective batch size of 8. Models were trained for three epochs without early stopping, and their final checkpoints were evaluated.

The causal language-modeling loss covered both prompt and response tokens, without completion-only masking. Training examples were capped at 4,096 tokens, without packing, and sequences exceeding the cap were truncated from the right; this could remove part of the reference response. Appendix C, Table C1 reports the training-length audit.

### 2.3. Experimental Design

The experimental design comprised four analyses addressing the main performance comparison between BERT and LLMs, the effects of output-format constraints and generation budgets, and differences across model configurations. All models were evaluated on the same 433 test sentences.

#### 2.3.1. Comparison of Zero-Shot and Fine-Tuned LLMs with BERT Baselines (RQ1)

We compared nine open-weight LLMs under zero-shot and task-specific fine-tuning conditions with two fine-tuned BERT baselines (UTH-BERT and NICT-BERT). All zero-shot LLMs were evaluated with a maximum output limit of 3,072 tokens. Fine-tuned LLMs were evaluated at 900 tokens and a larger limit of 4,096 tokens, except for the two LLM-jp-3.1 models, which used 2,600 tokens as their larger limit because of context-length restrictions. These comparisons assessed the extraction performance of the evaluated zero-shot and fine-tuned configurations relative to the supervised BERT baselines. Two supplementary reference models, Qwen3.6-35B and GPT-OSS-120B, were evaluated only in zero-shot settings (Appendix E, Table E1).

### 2.3.2. Effects of Output-format Constraints (RQ2)

To examine the effects of output-format constraints, all nine LLMs were evaluated under standard zero-shot prompting (ZS) and JSON Schema-constrained decoding (ZS+C), using the same prompt and a maximum of 3,072 output tokens. In ZS+C, XGrammar through vLLM enforced a JSON array of objects containing label and text, prohibited additional properties, and restricted labels to the fixed 109-category inventory (Kwon et al., 2023; Dong et al., 2024). Optional whitespace outside JSON strings was disabled. Supplementary constrained-decoding format diagnostics are provided in Appendix B.

Because LLM-jp-4-32B and SIP-jmed-13B generate explicit reasoning text before producing their final answers, applying JSON Schema constraints to the entire response also restricted the format of this intermediate output. We therefore introduced a supplementary two-stage condition (ZS+T+C), in which unconstrained reasoning generated in the original ZS condition was reused, and JSON Schema constraints were applied only to the subsequent answer-generation stage, with an additional allowance of 900 output tokens. Because total generation was not budget-matched to the single-stage conditions, this was treated as a supplementary diagnostic comparison.

### 2.3.3. Effects of Generation Budget (RQ3)

The nine fine-tuned LLMs were evaluated at 900 tokens and at the larger model-specific output limits described above. All three seeds were evaluated with the same checkpoint at each limit. To examine the budget–performance relationship in greater detail, seed-42 checkpoints were additionally evaluated at 512, 1,024, and 2,048 tokens, producing five budget settings per model. The seven models with larger contexts used 4,096 tokens as the upper limit; the two LLM-jp-3.1 models used 2,600 tokens under their 4,096-token context limit. These comparisons were designed to assess how output-token limits influenced extraction performance and the completeness of generated responses.

### 2.3.4. Comparison of Medical and General-purpose LLMs (RQ4)

We examined three medical–general pairs: SIP-jmed-13B versus LLM-jp-3.1-13B, MedGemma-27B versus Gemma-3-27B, and Med42-70B versus Llama-3.3-70B. Each pair was compared in ZS and ZS+C at 3,072 tokens and, after fine-tuning, at 900 tokens and matched larger limits for each seed. The larger limit was 2,600 tokens for both members of the 13B pair (including an additional matched run of SIP-jmed-13B) and 4,096 for the 27B and 70B pairs. Model size and architecture were examined descriptively.

## 2.4. Output Processing and Evaluation

### 2.4.1. Prediction Extraction and JSON Processing

LLM-generated responses were parsed to obtain entity predictions in the requested JSON format. For zero-shot inference, the last complete JSON array was selected; for fine-tuned inference, the first complete array was used. For Harmony-formatted responses, predictions were extracted from the final-answer channel. When a JSON array was truncated, complete entity objects were retained and the remaining structure was closed where possible; responses without a recoverable array were treated as empty predictions. After normalization, LLM predictions with empty text were discarded, while predictions with labels outside the prompt inventory or text not appearing verbatim in the sentence were retained. Output-integrity categories distinguished valid arrays, extra surrounding text, recoverable truncation, and parsing failures (Appendix D, Table D1).

### 2.4.2. Entity Matching and Performance Metrics

To compare BERT and LLM systems consistently, predictions were evaluated as sentence-level label–surface pairs rather than character-offset spans. BERT spans were converted into labeled surface strings, whereas LLM predictions were read from their label and text fields. Unicode NFKC normalization and removal of leading and trailing whitespace were applied to both references and predictions, followed by deduplication of identical label–surface pairs within each sentence. This reduced 15,119 annotated test mentions to 12,919 unique reference pairs.

Exact-surface matching required identical labels and normalized surface strings. Relaxed matching required identical labels and accepted a match if either normalized surface string was a substring of the other. A deterministic greedy one-to-one procedure processed predictions in output order. For each prediction, the first still-unmatched reference with an identical label and surface form was preferred; otherwise, the first still-unmatched reference satisfying the applicable criterion was selected in original reference order. Each prediction and reference could be matched at most once.

Micro-averaged precision, recall, and F1 were calculated from aggregate true-positive, false-positive, and false-negative counts. Macro48 was defined as mean per-category F1 over the 48 categories with at least 20 unique test reference pairs; categories with references but no correct predictions contributed an F1 of zero. LLMs used 109 available prompt labels, UTH-BERT 112, and NICT-BERT 113. The LLM inventory originated from the historical UTH-BERT mapping; each BERT inventory reflected its corresponding CoNLL conversion. Gold references were not restricted to model-specific label inventories, and the single test reference outside the LLM/UTH-BERT inventories (PN-None-others) counted as a false negative if unmatched.

A supplementary evaluation retained repeated occurrences of the same label–surface pair in both references and predictions, using the same matching criteria without character-offset alignment (Supplementary Table S1).

### 2.5. Statistical Analysis

Fine-tuned performance was summarized as mean ± standard deviation over the three seeds (42, 1, 2). Test-sample uncertainty was assessed using 2,000 bootstrap replicates at the document level, sampling the 36 test documents with replacement while keeping constituent sentences together. Performance metrics and paired differences were recalculated using the same resampled documents for each compared system, yielding 95% confidence intervals.

Bootstrap confidence intervals for the main BERT–LLM comparisons used seed-42 predictions, while medical–general comparisons were evaluated separately by seed.

### 2.6. Implementation and Reproducibility

Training was implemented with PyTorch and Hugging Face Transformers; QLoRA used TRL, PEFT, and bitsandbytes. Main-model inference used vLLM with XGrammar for constrained decoding; supplementary reference models used Ollama. All main experiments ran on local GPU resources, with 4,096-token inference using two to four 48-GB GPUs. Model identifiers, software versions, and detailed configurations are listed in Appendix F.

Source code, inference configurations, model identifiers, and derived evaluation files will be made publicly available in a code repository.

## 3. Results

All systems were evaluated on the same 433 test sentences. Fine-tuned performance is reported as the mean ± standard deviation across three training seeds unless otherwise specified. Bootstrap confidence intervals describe the designated fixed-seed predictions.

### 3.1. Overall Extraction Performance (RQ1)

The two fine-tuned BERT baselines achieved relaxed F1 scores of 0.836 (UTH-BERT) and 0.841 (NICT-BERT), with corresponding exact-surface F1 scores of 0.764 and 0.803 (Table 2). Under the common 3,072-token zero-shot condition, all nine main LLMs performed below the BERT baselines. MedGemma-27B achieved the highest zero-shot scores (relaxed F1, 0.337; exact-surface F1, 0.218). The two supplementary zero-shot reference models, Qwen3.6-35B and GPT-OSS-120B, achieved relaxed F1 values of 0.300 and 0.265, respectively.

At the larger fine-tuned generation limits, all eight LLMs with at least 13B total parameters exceeded both BERT baselines in three-seed mean F1 under both criteria (Table 2; Figure 2). Relaxed F1 ranged from 0.882 to 0.914 and exact-surface F1 from 0.840 to 0.880. Llama-3.3-70B achieved the highest mean under both criteria. The 1.8B LLM achieved 0.833 relaxed F1 (below both BERT means) and 0.780 exact-surface F1 (between the baselines).

In the paired seed-42 document-bootstrap analysis, relaxed-F1 differences versus UTH-BERT ranged from +0.048 (95% CI +0.026 to +0.078) for LLM-jp-3.1-13B to +0.079 (+0.055 to +0.114) for Gemma-3-27B among the eight larger LLMs. For the 1.8B model, the difference was approximately zero (95% CI −0.024 to +0.029). Additional precision, recall, and Macro48 metrics are reported in Supplementary Table S2.

Table 2. Overall BERT and LLM extraction performance under zero-shot and larger-budget fine-tuned configurations.

| **Model** | **Zero-shot relaxed F1 [CI]** | **Zero-shot exact F1 [CI]** | **FT relaxed mean ± SD [CI]** | **FT exact mean ± SD [CI]** | **FT relaxed Δ vs UTH [CI]** |
|---|---|---|---|---|---|
| UTH-BERT | — | — | 0.836 ± 0.004 [0.794–0.863] | 0.764 ± 0.005 [0.722–0.789] | — |
| NICT-BERT | — | — | 0.841 ± 0.001 [0.806–0.870] | 0.803 ± 0.002 [0.767–0.837] | +0.009 [+0.002, +0.015] |
| Llama-3.3-70B | 0.273 [0.245–0.299] | 0.159 [0.139–0.178] | 0.914 ± 0.004 [0.900–0.922] | 0.880 ± 0.006 [0.858–0.887] | +0.079 [+0.054, +0.113] |
| Med42-70B | 0.136 [0.119–0.152] | 0.043 [0.038–0.048] | 0.911 ± 0.003 [0.893–0.921] | 0.876 ± 0.005 [0.854–0.887] | +0.075 [+0.055, +0.105] |
| MedGemma-27B | 0.337 [0.317–0.356] | 0.218 [0.197–0.239] | 0.908 ± 0.006 [0.887–0.915] | 0.871 ± 0.008 [0.844–0.878] | +0.069 [+0.049, +0.100] |
| Gemma-3-27B | 0.319 [0.298–0.339] | 0.167 [0.149–0.183] | 0.907 ± 0.004 [0.900–0.923] | 0.871 ± 0.004 [0.861–0.890] | +0.079 [+0.055, +0.114] |
| SIP-jmed-13B | 0.025 [0.020–0.030] | 0.014 [0.011–0.018] | 0.899 ± 0.003 [0.881–0.911] | 0.859 ± 0.001 [0.840–0.876] | +0.064 [+0.041, +0.096] |
| Qwen3-30B-A3B | 0.235 [0.217–0.254] | 0.139 [0.124–0.154] | 0.894 ± 0.006 [0.868–0.901] | 0.852 ± 0.007 [0.823–0.861] | +0.054 [+0.030, +0.088] |
| LLM-jp-4-32B | 0.150 [0.133–0.164] | 0.068 [0.059–0.077] | 0.890 ± 0.006 [0.867–0.898] | 0.847 ± 0.008 [0.818–0.855] | +0.051 [+0.029, +0.083] |
| LLM-jp-3.1-13B | 0.038 [0.034–0.043] | 0.013 [0.011–0.015] | 0.882 ± 0.002 [0.857–0.901] | 0.840 ± 0.003 [0.809–0.861] | +0.048 [+0.026, +0.078] |
| LLM-jp-3.1-1.8B | 0.017 [0.014–0.019] | 0.001 [0.001–0.002] | 0.833 ± 0.003 [0.807–0.856] | 0.780 ± 0.003 [0.751–0.804] | +0.000 [−0.024, +0.029] |
| Qwen3.6-35B | 0.300 [0.281–0.321] | 0.153 [0.135–0.173] | — | — | — |
| GPT-OSS-120B | 0.265<br>[0.240–0.288] | 0.131<br>[0.115–0.147] | — | — | — |

*Zero-shot CIs are document-bootstrap 95% intervals for individual runs. FT scores are means ± SD across three seeds, but the bracketed FT CIs and paired Δ intervals are for fixed seed 42, not CIs for the three-seed means. FT-large is 4,096 output tokens except for the two LLM-jp-3.1 models (2,600); BERT and supplementary reference models have no FT output generation condition. Dashes indicate conditions not evaluated. Reference models used Ollama rather than the main vLLM inference pipeline.*

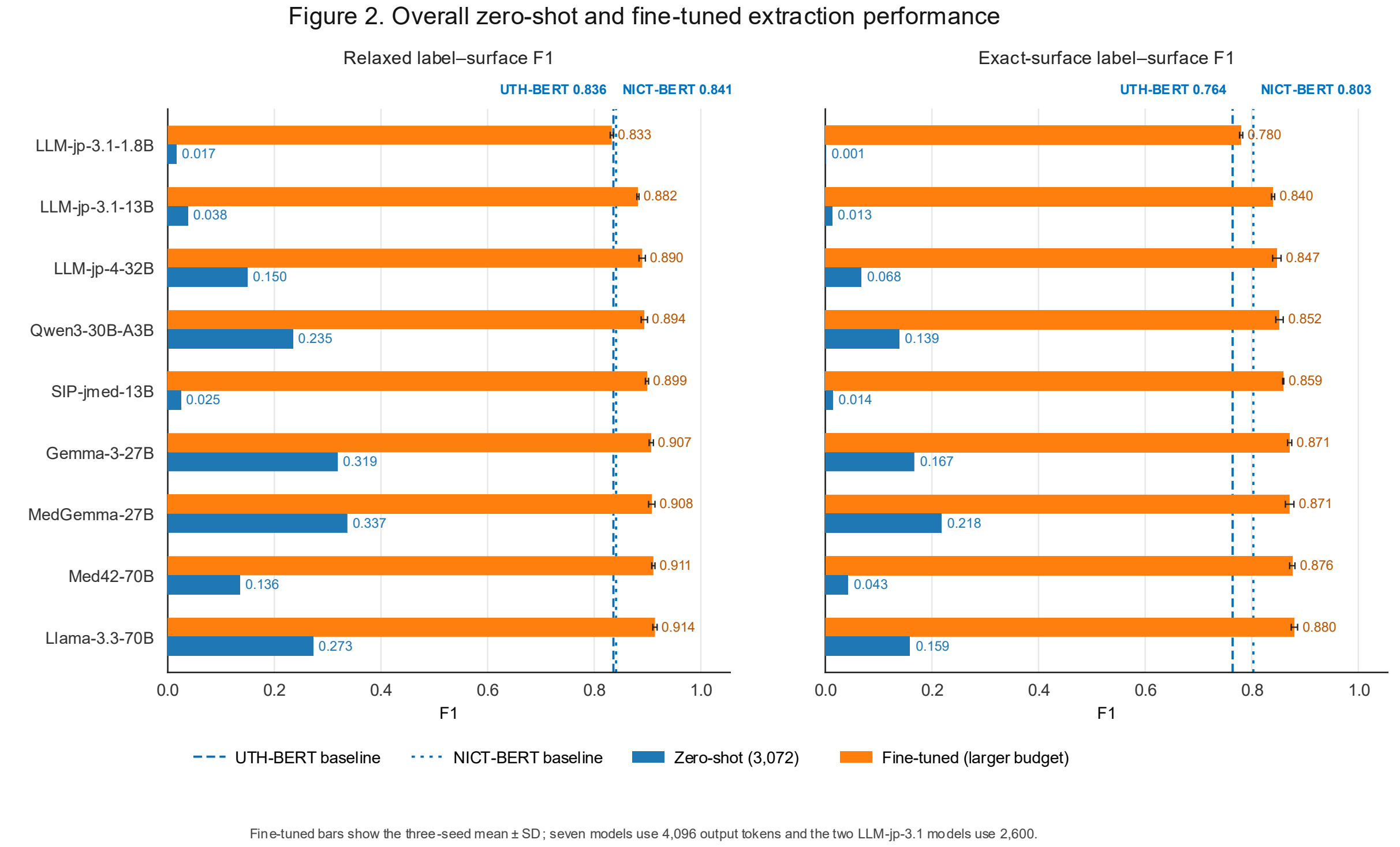

Figure 2. Relaxed and exact-surface F1 of the nine main LLMs under zero-shot prompting and larger-budget fine-tuning. Fine-tuned scores are three-seed means with SD error bars; dashed reference lines indicate the two BERT baselines. The two LLM-jp-3.1 models used 2,600 rather than 4,096 output tokens in the FT-large condition.

## 3.2. Effects of Output-format Constraints (RQ2)

For the seven models other than LLM-jp-4-32B and SIP-jmed-13B, the change in relaxed F1 from ZS to ZS+C ranged from −0.014 to +0.008 (Table 3). MedGemma-27B decreased from 0.337 to 0.323 and Gemma-3-27B from 0.319 to 0.308, while Qwen3-30B-A3B increased from 0.235 to 0.243. No constrained condition exceeded the best prompt-only zero-shot result. Additional zero-shot output diagnostics are provided in Supplementary Table S3.

The two reasoning-oriented models showed larger changes when the whole output was constrained. LLM-jp-4-32B declined from 0.150 / 0.068 to 0.000 / 0.000 (relaxed / exact-surface F1), and SIP-jmed-13B declined from 0.025 / 0.014 to 0.016 / 0.003. Under the supplementary ZS+T+C condition, their scores were 0.172 / 0.081 and 0.048 / 0.023, respectively, corresponding to relaxed-F1 increases of 0.022 and 0.023 relative to ZS; both remained below the strongest prompt-only model.

Table 3. Zero-shot extraction performance with and without JSON Schema constraints (RQ2).

| Model | ZS relaxed / exact | ZS+C relaxed / exact | Δ relaxed (ZS+C − ZS) | ZS+T+C relaxed / exact |
|---|---|---|---|---|
| Llama-3.3-70B | 0.273 / 0.159 | 0.270 / 0.160 | −0.003 | — |
| Med42-70B | 0.136 / 0.043 | 0.143 / 0.044 | +0.007 | — |
| MedGemma-27B | 0.337 / 0.218 | 0.323 / 0.207 | −0.014 | — |
| Gemma-3-27B | 0.319 / 0.167 | 0.308 / 0.158 | −0.011 | — |
| SIP-jmed-13B | 0.025 / 0.014 | 0.016 / 0.003 | −0.009 | 0.048 / 0.023 |
| Qwen3-30B-A3B | 0.235 / 0.139 | 0.243 / 0.141 | +0.008 | — |
| LLM-jp-4-32B | 0.150 / 0.068 | 0.000 / 0.000 | −0.150 | 0.172 / 0.081 |
| LLM-jp-3.1-13B | 0.038 / 0.013 | 0.035 / 0.012 | −0.003 | — |
| LLM-jp-3.1-1.8B | 0.017 / 0.001 | 0.013 / 0.001 | −0.003 | — |

*ZS and ZS+C used 3,072 output tokens. ZS+T+C was evaluated only for SIP-jmed-13B and LLM-jp-4-32B and included a separate additional 900-token answer-generation allowance; it was not matched on total generation budget.*

## 3.3. Effects of Generation Budget (RQ3)

At the 900-token limit, the eight fine-tuned LLMs with at least 13B parameters achieved three-seed mean relaxed F1 scores of 0.817–0.830, below the BERT baseline means. The seed-42 differences versus UTH-BERT were small, and the paired confidence intervals included zero. The 1.8B LLM achieved 0.781. Increasing to the larger model-specific generation budgets raised relaxed F1 for all nine models (Table 4).

Across the three-seed means, relaxed F1 increased by 0.052–0.085 from FT-900 to FT-large. For seed 42, the corresponding improvements were 0.053–0.086. Recall increased with the output limit, while precision changed relatively little. At 900 tokens, 79–87 sentences per model reached the output limit, accounting for approximately 50%–53% of reference entities. For models with at least 13B parameters, recall on truncated sentences was 0.57–0.59, versus 0.91–0.94 on other sentences (Supplementary Table S4).

At the larger budgets, two or three sentences reached the output limit for models using 4,096 tokens, compared with six and ten for the two LLM-jp-3.1 models using 2,600 tokens. The five-point seed-42 analysis showed increasing relaxed F1 at every successive budget, from 0.676–0.716 at 512 tokens to 0.830–0.897 at 2,048 tokens (Figure 3; Supplementary Table S5). For Llama-3.3-70B, F1 increased from 0.894 at 2,048 tokens to 0.911 at 4,096, while the number of sentences reaching the limit fell from 11 to two.

Table 4. Fine-tuned extraction performance at 900 and larger output-token budgets (RQ3).

| Model | FT-900 relaxed / exact | FT-large relaxed / exact | Δ relaxed mean | Length stops 900 → large | FT-large limit |
|---|---|---|---|---|---|
| Llama-3.3-70B | 0.829 ± 0.003 | 0.914 ± 0.004 | +0.085 | 86 → 2 | 4,096 |
| | 0.800 ± 0.006 | 0.880 ± 0.006 | | | |
| Med42-70B | 0.829 ± 0.002 | 0.911 ± 0.003 | +0.081 | 87 → 2 | 4,096 |
| | 0.800 ± 0.004 | 0.876 ± 0.005 | | | |
| MedGemma-27B | 0.830 ± 0.003 | 0.908 ± 0.006 | +0.078 | 81 → 2 | 4,096 |
| | 0.798 ± 0.004 | 0.871 ± 0.008 | | | |
| Gemma-3-27B | 0.830 ± 0.002 | 0.907 ± 0.004 | +0.078 | 84 → 2 | 4,096 |
| | 0.799 ± 0.002 | 0.871 ± 0.004 | | | |
| SIP-jmed-13B | 0.822 ± 0.002 | 0.899 ± 0.003 | +0.078 | 84 → 3 | 4,096 |
| | 0.788 ± 0.002 | 0.859 ± 0.001 | | | |
| Qwen3-30B-A3B | 0.817 ± 0.002 | 0.894 ± 0.006 | +0.076 | 85 → 3 | 4,096 |
| | 0.782 ± 0.004 | 0.852 ± 0.007 | | | |
| LLM-jp-4-32B | 0.820 ± 0.004 | 0.890 ± 0.006 | +0.071 | 79 → 2 | 4,096 |
| | 0.782 ± 0.006 | 0.847 ± 0.008 | | | |
| LLM-jp-3.1-13B | 0.819 ± 0.001 | 0.882 ± 0.002 | +0.063 | 82 → 6 | 2,600 |
| | 0.782 ± 0.002 | 0.840 ± 0.003 | | | |
| LLM-jp-3.1-1.8B | 0.781 ± 0.002 | 0.833 ± 0.003 | +0.052 | 81 → 10 | 2,600 |
| | 0.734 ± 0.002 | 0.780 ± 0.003 | | | |

*F1 entries are three-seed mean ± SD, displayed as relaxed (top) and exact-surface (bottom). Δ relaxed is the difference between corresponding three-seed means. Length-stop counts are from seed 42, not three-seed means.*

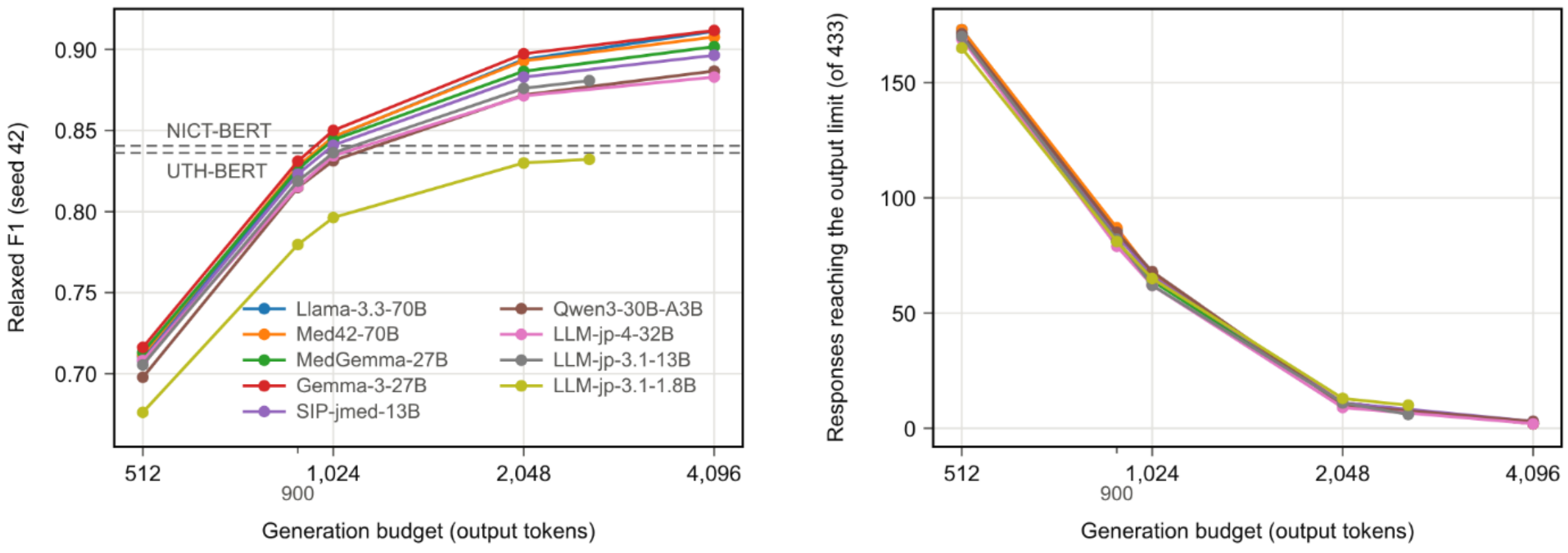

Figure 3. Effect of generation-token limits on relaxed F1 and the number of sentences reaching the output limit (seed 42). Left: relaxed F1; right: number of sentences reaching the configured output limit. The two LLM-jp-3.1 models ended at 2,600 tokens; the other seven models ended at 4,096.

## 3.4. Model Configurations and Medical Specialization (RQ4)

At the larger FT budgets, the two dense 27B models, Gemma-3-27B and MedGemma-27B, achieved mean relaxed F1 scores of 0.907 and 0.908, respectively, compared with 0.914 and 0.911 for Llama-3.3-70B and Med42-70B. The two MoE models, Qwen3-30B-A3B and LLM-jp-4-32B, achieved 0.894 and 0.890, respectively (Table 2).

In zero-shot evaluation, the medical-minus-general relaxed-F1 differences were +0.018 for MedGemma versus Gemma, −0.013 for SIP-jmed versus LLM-jp-3.1-13B, and −0.137 for Med42 versus Llama; the directions did not change under ZS+C (Table 5).

After fine-tuning under matched larger budgets, SIP-jmed-13B achieved relaxed F1 of approximately 0.890 at 2,600 tokens, compared with 0.882 for LLM-jp-3.1-13B, and positive differences across the three seeds (+0.007 to +0.008). MedGemma–Gemma differences changed direction across seeds (−0.010 to +0.007), whereas Med42 was slightly below Llama in each seed (approximately −0.004). At 900 tokens, the three-seed mean relaxed differences across all three pairs were close to zero (Table 5). Seed-specific paired bootstrap comparisons are retained in Supplementary Table S6.

Table 5. Medical-minus-general differences in relaxed and exact-surface F1 across zero-shot and matched fine-tuned conditions (RQ4).

| **Medical − general pair** | **ZS**<br>**Δrelaxed / Δexact** | **ZS+C**<br>**Δrelaxed / Δexact** | **FT-900**<br>**Δrelaxed / Δexact** | **FT-large**<br>**Δrelaxed / Δexact** |
|---|---|---|---|---|
| SIP-jmed − LLM-jp-3.1 | −0.013 / +0.001 | −0.020 / −0.009 | +0.002 / +0.005 | +0.007 / +0.010 |
| MedGemma − Gemma | +0.018 / +0.052 | +0.015 / +0.048 | +0.000 / −0.001 | +0.001 / −0.001 |
| Med42 − Llama | −0.137 / −0.115 | −0.127 / −0.116 | +0.000 / +0.000 | −0.004 / −0.004 |

*Positive Δ favors the medical model. ZS results are single runs; FT differences are based on three-seed mean F1. The 13B pair uses 2,600-token matched FT-large runs, including SIP-jmed-13B at 2,600 tokens, while 27B and 70B pairs use 4,096. Seed-specific paired bootstrap 95% CIs (relaxed F1) appear in Supplementary Table S6.*

## 3.5. Supplementary Evaluation and Label-level Performance

Retaining repeated entity mentions produced results consistent with the primary deduplicated analysis (Supplementary Table S1). For the eight fine-tuned LLMs with at least 13B parameters, relaxed F1 ranged from 0.882 to 0.918, changing by approximately 0.000 to +0.004 relative to the primary analysis. UTH-BERT and NICT-BERT increased to 0.845 and 0.847, respectively; the 1.8B LLM decreased to 0.819. At seed 42, Macro48 ranged from 0.832 to 0.872 for the eight larger LLMs, compared with 0.808 for UTH-BERT and 0.803 for NICT-BERT (Supplementary Table S2).

Label-level performance varied across categories. At seed 42 with 4,096 output tokens, Med42-70B exceeded UTH-BERT on state-patient (relaxed F1 0.91 vs. 0.79), clinical_test (0.93 vs. 0.81), treatment (0.91 vs. 0.85), and drug (0.95 vs. 0.81), as well as PN-Negative-patient (0.95 vs. 0.94) and PN-Negative-others (0.92 vs. 0.82). Both systems performed less well on state-others (0.56 vs. 0.45) and value-others (0.28 vs. 0.29). Detailed results are given in Supplementary Tables S7–S8 and Supplementary Figure S1.

# 4. Discussion

## 4.1. Task adaptation and the BERT–LLM comparison

The principal finding of this study is that the relative performance of BERT-based and generative LLM approaches to fine-grained Japanese clinical NER varied substantially with task-specific adaptation and generation budget. Under zero-shot prompting, all evaluated LLMs performed well below the supervised BERT baselines, with the best relaxed F1 reaching only 0.337. In contrast, after task-specific fine-tuning and with larger generation budgets, all eight LLMs with at least 13B total parameters achieved higher mean relaxed and exact-surface F1 scores than both BERT baselines. The 1.8B model remained below both baselines in relaxed F1, indicating that this advantage was not observed uniformly across all model configurations.

These findings are consistent with previous research emphasizing the importance of task-specific adaptation in clinical and biomedical NER. Naguib et al. (2024) reported that prompt-based generative models underperformed supervised encoders in multilingual clinical NER, whereas Keloth et al. (2024) demonstrated that instruction tuning could narrow the performance gap between general-purpose LLMs and specialized biomedical encoders. In Japanese clinical NLP, Shimizu et al. (2025) reported that a fine-tuned LLM outperformed BERT for disease name recognition. Our results extend these observations to a broader annotation scheme comprising 113 clinical entity categories. The large difference between zero-shot and fine-tuned configurations suggests that task-specific supervision is particularly relevant when models must follow detailed annotation conventions and distinguish numerous closely related entity categories. However, because the zero-shot and fine-tuned conditions also differed in their inference settings, this comparison should not be interpreted as an isolated estimate of the effect of fine-tuning.

The supplementary zero-shot evaluations of Qwen3.6-35B and GPT-OSS-120B showed similarly limited performance relative to the supervised baselines. Although these models were evaluated using a different inference engine, their results provide additional evidence that the observed zero-shot performance gap was not confined to the nine main models.

## 4.2. Output constraints and reasoning-aware generation

The limited zero-shot performance raised the question of whether difficulties in producing the required output format contributed to extraction errors. For the seven models without the two supplementary reasoning-aware conditions, JSON Schema-constrained decoding produced only small changes in relaxed F1 (−0.014 to +0.008). Although the constraints enforced valid output structure and restricted labels to the predefined inventory, they did not substantially improve extraction accuracy. This suggests that output-format enforcement alone was insufficient to address the difficulties of zero-shot fine-grained clinical NER.

The two reasoning-oriented models exhibited a different response to constrained decoding. Applying JSON Schema constraints to the entire generated response substantially reduced extraction performance, with LLM-jp-4-32B producing no correctly matched entities under this condition. The supplementary two-stage approach, which reused unconstrained reasoning and applied constraints only to the final answer, improved performance relative to the original zero-shot condition for both models. Nevertheless, these improvements were modest, and the resulting scores remained below those of the strongest zero-shot models. Because the two-stage condition included an additional answer-generation allowance, the observed differences cannot be attributed solely to separating reasoning from constrained output generation. These findings suggest that the interaction between explicit reasoning, output constraints, and generation capacity warrants consideration when applying structured decoding to reasoning-oriented LLMs.

## 4.3. Generation budgets and extraction completeness

Generation budget had a substantial influence on the observed performance of fine-tuned LLMs. At the 900-token output limit, the eight models with at least 13B total parameters achieved mean relaxed F1 scores below both BERT baselines. At the larger model-specific limits, all eight exceeded the BERT baseline scores. For identical seed-42 checkpoints, increasing the output allowance improved relaxed F1 by 0.053–0.086 across the nine LLMs. Thus, the relative performance of BERT and LLM systems depended not only on the models being compared but also on the output capacity available during inference.

The improvement was closely associated with reduced output truncation and increased recall. At the 900-token limit, approximately one-fifth of the test sentences reached the generation limit, yet these sentences contained approximately half of the reference entities. Recall was considerably lower on these sentences than on responses that completed within the limit. Increasing the output allowance substantially reduced truncation and improved recall, while precision changed relatively little. These observations support output incompleteness as an important contributor to the lower F1 scores under restrictive generation limits. However, sentences reaching the output limit also tended to contain more entities, and the observed recall differences should not be attributed exclusively to truncation without accounting for sentence complexity.

These findings have implications for the evaluation of generative NER systems. Unlike sequence-labeling models, which predict entity labels over the input sequence, generative models must allocate output tokens to represent the extracted entities. In entity-dense text, an insufficient generation allowance can therefore affect measured extraction performance independently of changes to model weights. Reporting the configured generation budget, context-length restrictions, and output truncation is consequently important for interpreting and reproducing comparisons involving generative extraction.

### 4.4. Model size and medical specialization

Model performance varied across parameter sizes and architectures, although larger models did not consistently achieve proportionally higher extraction accuracy. After fine-tuning, the 27B dense models achieved relaxed F1 scores close to those of the 70B models, while the two MoE models also exceeded the BERT baselines. In contrast, the 1.8B model remained below both baselines in relaxed F1. These findings indicate that, among the evaluated configurations, LLMs smaller than 70B could achieve competitive fine-grained clinical NER performance.

Medical specialization did not provide a consistent advantage across the three selected model pairs. At matched larger generation budgets, SIP-jmed-13B achieved modestly higher relaxed F1 than LLM-jp-3.1-13B across all three training seeds. In contrast, MedGemma-27B and Gemma-3-27B showed small differences that changed direction across seeds, while Med42-70B scored slightly below Llama-3.3-70B. The zero-shot comparisons similarly showed no uniform advantage for the medically specialized models. These findings suggest that medical-domain specialization alone is not a reliable indicator of superior performance on this task.

### 4.5. Practical implications and residual errors

Although the larger fine-tuned LLMs achieved higher overall F1 scores than the BERT baselines, their advantages were not uniform across entity categories. Med42-70B performed better than UTH-BERT on several frequent categories, including state-patient, clinical_test, treatment, and drug, as well as the two evaluated PN-Negative categories. In contrast, both approaches showed relatively low F1 for state-others (0.56 versus 0.45) and value-others (0.28 versus 0.29). For change-related categories, Med42-70B was below UTH-BERT on quantity_progress-increase-patient (0.67 versus 0.81) and quality_progress-improve-patient (0.87 versus 0.94). These differences highlight the need to examine category-level performance rather than aggregate F1 alone (Supplementary Figure S1).

Closer inspection of the seed-42 Med42-70B predictions revealed residual difficulties with less frequent, context-dependent labels. Recall was 0.41 for state-others (155 reference pairs) and 0.22 for value-others (63), versus 0.92 for state-patient (2,318) and 0.94 for value-patient (921). In contrast, PN-Negative categories were recognized relatively well (relaxed F1: 0.95 for patient and 0.92 for others). In one test sentence, the numeral

“3” in an HbA1c conversion expression was annotated as value-others but predicted as value-patient. These contrasts suggest sensitivity to fine-grained annotation conventions, although category frequency and the specifics of individual annotations may contribute. The selected recalls and example are given in Supplementary Tables S8 and S9.

Entity-boundary fragmentation was another observed error. The reference annotated “真皮浅層” as a single tissue-patient entity, whereas Med42-70B predicted “真皮” as tissue-patient and “浅層” as spacial_concept. Because relaxed matching accepts substring overlap when the labels agree, a partial extraction can be credited even when the original entity boundary is not reproduced. The difference between Med42-70B relaxed and exact-surface F1 (0.908 versus 0.871, seed 42) further motivates reporting both criteria, although this example does not quantify how much of the difference is attributable to fragmentation. The example is documented in Supplementary Table S9.

Distinguishing who experiences a finding and correctly recognizing changes in clinical condition remain important for clinical information extraction, because such errors can alter the interpretation of clinical narratives. The relatively strong PN-Negative results indicate that the observed residual difficulties should not be generalized to negation extraction. Nevertheless, improvements in overall extraction performance do not imply comparable reliability across all clinically important entity categories.

The experiments also demonstrate that open-weight LLMs can be fine-tuned and evaluated locally for this fine-grained extraction task. Several models substantially smaller than 70B achieved competitive accuracy, suggesting that the largest evaluated models are not always necessary to obtain strong extraction performance. However, model selection for practical applications should consider category-specific errors alongside computational requirements and deployment constraints. Further validation on routine clinical records would be necessary before extending these findings to operational healthcare settings.

# 5. Conclusion

This study compared BERT-based models and open-weight LLMs for fine-grained Japanese clinical NER, examining task-specific fine-tuning, output-format constraints, generation budgets, and model configurations. Zero-shot LLMs performed substantially below the supervised BERT baselines, whereas all eight fine-tuned LLMs with at least 13B total parameters achieved higher mean F1 scores at larger generation budgets. Increasing the output allowance for identical fine-tuned checkpoints reduced truncation and improved recall, substantially affecting the relative performance of LLMs and BERT. In contrast, JSON Schema constraints alone provided limited improvements in zero-shot extraction, and medically specialized models showed no consistent advantage over their general-purpose comparators. These findings highlight the importance of task-specific adaptation and adequate generation capacity in generative clinical NER, while emphasizing the need to account for inference settings and category-specific errors when comparing extraction systems.

## Limitations

This study has several limitations. First, published case reports may differ from routine electronic health records in their structure, terminology, and documentation practices. Evaluation on additional datasets and clinical settings is therefore needed.

Second, the primary evaluation used sentence-level label–surface matching rather than exact character-span matching. Identical label–surface pairs were deduplicated within each sentence, and relaxed matching accepted partial surface overlap when entity labels agreed. Although the supplementary evaluation retaining repeated mentions produced similar overall findings, neither evaluation directly assessed the localization of entity boundaries in the original text. Consequently, the reported F1 scores should not be interpreted as conventional span-level NER accuracy.

Third, the experiments involved differences in model architectures, training procedures, and inference configurations. The LLM fine-tuning procedure applied the training loss to both prompt and response tokens, and sequences exceeding the training context limit were truncated. In addition, the fixed prompt label inventory was derived from a historical annotation mapping and did not exactly match the labels present in the current training partition, although the affected categories were absent from the test references. These implementation choices may influence performance, and their individual effects were not isolated in the present experiments. Likewise, differences between zero-shot and fine-tuned configurations cannot be attributed solely to task adaptation because their generation budgets were not identical. The unchanged zero-shot prompt instructed models not to output the literal "なし", although that surface form occurred in 56 test reference pairs, potentially affecting recall for those particular mentions.

Finally, the medical–general comparisons were also limited to three selected model pairs and cannot establish the independent effect of domain specialization. Further studies incorporating additional datasets, training configurations, and controlled model comparisons would help assess the robustness and broader applicability of the findings.

## Declarations

## Author contributions

Conceptualization—Feng Han and Ayako Yagahara; methodology, software, experiments and analysis—Feng Han; resources—Kenji Hirata, Ayako Yagahara and Kohsuke Kudo; original drafting—Feng Han; review and editing—Feng Han, Ayako Yagahara, Kenji Hirata, Atsushi Yoshida, and Kohsuke Kudo; supervision—Kohsuke Kudo.

## Funding

This research was supported by the Japan Agency for Medical Research and Development (AMED) under Grant Number JP266f0137006.

This research was supported by a research grant from the Telecommunications Advancement Foundation.

## Ethics and consent

The study analyzed a research-distributed corpus derived from previously published case reports.

## Conflict of Interest

The authors declare no conflicts of interest.

## Data and code availability

The iCorpus dataset must be obtained from its publisher under the applicable terms of use. Code, summary metrics, model identifiers and reproducibility files will be made publicly available in a code repository, subject to the underlying data-distribution restrictions.

## Appendix A. Japanese prompt template and label inventory

The following Japanese prompt is preserved verbatim as the model input; translating it would not reproduce the evaluated condition. The {labels} placeholder was replaced by the fixed 109-label inventory derived from the historical UTH-BERT NER_tag2idx_0.json mapping, one per line, and {text} by the input sentence.

The 109 labels are not the complete label inventory of the current training partition (111 categories). Three categories represented by one training entity each were absent from the prompt inventory but remained in the fine-tuning target responses: judge-patient, quantity_evaluation-normal-others, and quantity_evaluation-others. Conversely, quantity_progress-no_change-others was listed in the prompt but absent from the current training partition. None of these four categories occurred in the test partition.

```
あなたは日本語の臨床テキストからの固有表現抽出(NER)を行う専門家です。
以下の臨床テキストから、指定されたラベルに該当する表現をすべて抽出してください。

# ラベル一覧
{labels}

# 抽出ルール
- テキスト中に出現した表現をそのまま抜き出すこと(言い換え・要約をしない)
- 文や節ではなく、最小の固有表現(単語〜短い句)を抜き出すこと
- テキスト中に出現したラベルだけを出力すること(空文字や「なし」は出力しない)
- 該当が無ければ空配列 [] を返すこと
- 出力は必ず次形式のJSON配列のみ。説明文を付けない:
[{"label": "ラベル名", "text": "抽出した表現"}, ...]

# 臨床テキスト
{text}

# 出力(JSON配列のみ)
```

Fixed 109-label inventory (original label strings):

| Label 1 | Label 2 | Label 3 |
|---|---|---|
| ChiefComplaint | incident | quality_progress-recover-patient |
| FamilyHistory | info | quality_progress-start-others |
| MedicationHistory | intention-NG | quality_progress-start-patient |

| Label 1 | Label 2 | Label 3 |
|---|---|---|
| PN-Negative-others | intention-OK | quality_progress-worsen-others |
| PN-Negative-patient | item | quality_progress-worsen-patient |
| PN-None-patient | job-others | quantity_evaluation-High-others |
| PN-Positive-others | job-patient | quantity_evaluation-High-patient |
| PN-Positive-patient | judge-Negative-others | quantity_evaluation-Low-others |
| PastHistory | judge-Negative-patient | quantity_evaluation-Low-patient |
| PresentCondition | judge-None-others | quantity_evaluation-normal-patient |
| PresentHistory | judge-None-patient | quantity_evaluation-patient |
| SocialHistory | judge-Positive-others | quantity_progress-decrease-others |
| action | judge-Positive-patient | quantity_progress-decrease-patient |
| activity-others | judge-Suspicious-others | quantity_progress-increase-others |
| activity-patient | judge-Suspicious-patient | quantity_progress-increase-patient |
| age-others | judge-future-others | quantity_progress-no_change-others |
| age-patient | judge-future-patient | quantity_progress-no_change-patient |
| allergy-others | laterality | quantity_progress-patient |
| allergy-patient | organization | sex-others |
| body-others | person-family | sex-patient |
| body-patient | person-others | smoking-current-others |
| clinical_test | person-patient | smoking-current-patient |
| consultation | pose | smoking-stop-patient |
| detail | quality_evaluation-mild-others | spacial_concept |
| device | quality_evaluation-mild-patient | state-others |
| drinking-current-others | quality_evaluation-moderate-patient | state-patient |
| drinking-current-patient | quality_evaluation-others | substance |
| drug | quality_evaluation-patient | time |
| execute | quality_evaluation-severe-others | time_span |
| explain | quality_evaluation-severe-patient | tissue-others |
| exposure | quality_progress-improve-others | tissue-patient |
| food | quality_progress-improve-patient | treatment |
| frequency | quality_progress-no_change-others | unit |
| function-others | quality_progress-no_change-patient | value-others |
| function-patient | quality_progress-others | value-patient |
| gene-others | quality_progress-patient | gene-patient |
| quality_progress-recover-others | | |

## Appendix B. Constrained-decoding format diagnostics

JSON Schema decoding required an array of objects with label and text fields, prohibited additional properties and restricted labels to the 109-category inventory. In the final constrained setting, optional whitespace outside strings was disabled. This decision was informed by preliminary permissive-whitespace diagnostics, which are not budget-matched to the comparisons reported in the Results.

In a preliminary permissive-whitespace run, LLM-jp-4-32B generated whitespace until reaching the output limit for 334 of 433 sentences (relaxed F1 approximately 0.005). In the same preliminary diagnostic series, SIP-jmed-13B and LLM-jp-3.1-1.8B encountered length termination in 142 and 426 sentences, respectively. These observations motivated restricting optional whitespace; they do not replace the common 3,072-token ZS and ZS+C experiments reported in the Results.

## Appendix C. Fine-tuning sequence-length audit

Serialized user and assistant messages were limited to 4,096 tokens without packing; examples exceeding this length were truncated from the right. The following audit describes the same 1,672 training examples. Estimated entity loss is a proportional estimate, not a direct count of omitted complete entities.

**Table C1. Tokenizer-dependent training-sequence length audit.**

| Model | Examples >4,096 | Training fraction | Maximum tokens | Estimated entity loss |
|---|---|---|---|---|
| Llama-3.3-70B | 17 | 1.02% | 9,510 | 2.53% |
| Med42-70B | 17 | 1.02% | 9,485 | 2.48% |
| MedGemma-27B | 14 | 0.84% | 9,048 | 2.25% |
| Gemma-3-27B | 14 | 0.84% | 9,048 | 2.25% |
| SIP-jmed-13B | 15 | 0.90% | 9,096 | 2.21% |
| Qwen3-30B-A3B | 17 | 1.02% | 9,576 | 2.62% |
| LLM-jp-4-32B | 15 | 0.90% | 9,079 | 2.27% |
| LLM-jp-3.1-13B | 15 | 0.90% | 9,096 | 2.21% |
| LLM-jp-3.1-1.8B | 15 | 0.90% | 9,096 | 2.21% |

## Appendix D. Output integrity and parsing diagnostics

For fine-tuned runs, outputs were categorized as containing a complete recoverable array, requiring deterministic truncation repair, or being unparseable. Length termination was counted separately because a response can reach its limit after producing a complete answer. Unparseable responses were scored as empty predictions. The following aggregate counts are from the audit of the fine-tuned runs and do not include zero-shot parser failures.

**Table D1. Aggregate fine-tuned output parsing audit.**

| Configuration | Outputs | Complete arrays | Repaired | Failures | Length stops |
|---|---|---|---|---|---|
| FT: 900-token limit (27 runs) | 11,691 | 9,449 | 2,233 | 9 | 2,249 |
| FT: larger budgets (27 runs) | 11,691 | 11,590 | 92 | 9 | 98 |
| SIP-jmed: 2,600-token matched (3 runs) | 1,299 | 1,276 | 22 | 1 | 21 |

Across 57 fine-tuned runs, 19 unparseable responses were counted as empty predictions in the recorded scoring files. The audit did not identify legitimate empty JSON arrays in those fine-tuned runs. Parsing of zero-shot runs followed a separate last-array rule, with final-channel extraction for Harmony-formatted responses.

## Appendix E. Supplementary zero-shot reference models

Two additional open-weight models were evaluated as zero-shot references under Ollama, with a 3,072-token output limit and an 8,192-token context limit. These models did not participate in local QLoRA fine-tuning. They provide supplementary evidence, not a replacement for the nine-model vLLM comparison.

**Table E1. Supplementary zero-shot reference results.**

| Model | Condition | Relaxed F1 | Exact-surface F1 | Length stops | Reported fail/empty |
|---|---|---|---|---|---|
| Qwen3.6-35B | Reasoning off; ZS | 0.3003 | 0.1535 | 3 | 0 |
| Qwen3.6-35B | Reasoning on; answer schema | 0.0217 | 0.0173 | 310 | 310 |
| GPT-OSS-120B | Reasoning on; ZS | 0.2652 | 0.1311 | 16 | 8 |
| GPT-OSS-120B | Reasoning on; answer schema | 0.2630 | 0.1351 | 11 | 11 |

For Qwen3.6-35B with explicit reasoning and constrained answers, 310 sentences exhausted the output budget during reasoning. GPT-OSS-120B gave similar relaxed F1 with and without answer constraints (0.265 versus 0.263). These auxiliary runs used a different inference engine from the main experiments.

## Appendix F. Checkpoint identifiers and implementation configuration

The following identifiers were taken from the training metadata and model-lookup records. Repository revisions, adapter hashes, and further environment details will be provided with the public code release. The outputs of supplementary Ollama models are identified by their local names and digests.

**Table F1. Model identifiers used in the study.**

| Model | Checkpoint / local identifier |
|---|---|
| UTH-BERT | UTH_BERT_BASE_512_MC_BPE_WWM_V25000_352K |
| NICT-BERT | NICT_BERT-base_JapaneseWikipedia_32K_BPE |

| Model | Checkpoint / local identifier |
|---|---|
| LLM-jp-3.1-1.8B | llm-jp/llm-jp-3.1-1.8b-instruct4 |
| LLM-jp-3.1-13B | llm-jp/llm-jp-3.1-13b-instruct4 |
| SIP-jmed-13B | SIP-med-LLM/SIP-jmed-llm-3-13b-OP-32k-R0.1 |
| Gemma-3-27B | google/gemma-3-27b-it (locally converted text weights) |
| MedGemma-27B | google/medgemma-27b-text-it |
| Qwen3-30B-A3B | Qwen/Qwen3-30B-A3B-Instruct-2507 |
| LLM-jp-4-32B | llm-jp/llm-jp-4-32b-a3b-thinking |
| Llama-3.3-70B | meta-llama/Llama-3.3-70B-Instruct |
| Med42-70B | m42-health/Llama3-Med42-70B |
| Qwen3.6-35B | ollama qwen3.6:35b; digest 07d35212591f |
| GPT-OSS-120B | ollama gpt-oss:120b; digest a951a23b46a1 |

QLoRA: 4-bit NF4 with double quantization; bfloat16 computation; LoRA r = 16, α = 32, dropout = 0.05, no bias adapters; target modules q_proj, k_proj, v_proj, o_proj, gate_proj, up_proj, and down_proj. Training used AdamW (adamw_torch), learning rate $2 \times 10^{-4}$, cosine scheduling, 3% warm-up, weight decay 0, per-device batch size 1, gradient accumulation 8, three epochs, no early stopping, non-reentrant gradient checkpointing, and seeds 42, 1, and 2. Loss was applied to the combined prompt and answer sequence.

Software: vLLM 0.8.5 (XGrammar structured output), transformers 4.51.3, torch 2.6.0+cu124, peft 0.14.0, trl 0.13.0, bitsandbytes 0.49.2; separate BERT environment with torch 2.0.1+cu117 and transformers 4.40.2; Ollama 0.30.11. The primary experiments used local GPUs with 48 GB memory each. Full machine specifications and file hashes will be provided with the public code release.

## Appendix G. Supplementary performance tables

**Supplementary Table S1. Primary (deduplicated) and mention-level F1 of the fine-tuned systems (three-seed means; larger output limits).**

| System | Relaxed, primary | Relaxed, mention-level | Exact-surface, primary | Exact-surface, mention-level |
|---|---|---|---|---|
| Llama-3.3-70B | 0.914 | 0.918 | 0.880 | 0.881 |
| Med42-70B | 0.911 | 0.913 | 0.876 | 0.876 |
| MedGemma-27B | 0.908 | 0.912 | 0.871 | 0.871 |
| Gemma-3-27B | 0.907 | 0.910 | 0.871 | 0.871 |
| SIP-jmed-13B | 0.899 | 0.901 | 0.859 | 0.858 |
| Qwen3-30B-A3B | 0.894 | 0.896 | 0.852 | 0.851 |
| LLM-jp-4-32B | 0.890 | 0.893 | 0.847 | 0.847 |
| LLM-jp-3.1-13B | 0.882 | 0.882 | 0.840 | 0.836 |
| LLM-jp-3.1-1.8B | 0.833 | 0.819 | 0.780 | 0.763 |
| NICT-BERT | 0.841 | 0.847 | 0.803 | 0.808 |
| UTH-BERT | 0.836 | 0.845 | 0.764 | 0.768 |

*Mention-level scoring counts each of the 15,119 annotated mentions and each occurrence in the output separately, using the same greedy matching; the primary evaluation collapses repeated label–surface pairs within a sentence (12,919 pairs).*

**Supplementary Table S2. Seed-42 metrics for fine-tuned systems at the larger output limits.**

| Model | Precision | Recall | Relaxed F1 95% CI | Macro48 |
|---|---|---|---|---|
| UTH-BERT | 0.843 | 0.822 | [0.794, 0.863] | 0.808 |
| NICT-BERT | 0.841 | 0.841 | [0.806, 0.870] | 0.803 |
| Llama-3.3-70B | 0.916 | 0.907 | [0.900, 0.922] | 0.872 |
| Med42-70B | 0.912 | 0.903 | [0.893, 0.921] | 0.860 |
| MedGemma-27B | 0.907 | 0.897 | [0.887, 0.915] | 0.857 |

| Model | Precision | Recall | Relaxed F1<br>95% CI | Macro48 |
|---|---|---|---|---|
| Gemma-3-27B | 0.914 | 0.909 | [0.900, 0.923] | 0.870 |
| SIP-jmed-13B | 0.907 | 0.886 | [0.881, 0.911] | 0.853 |
| Qwen3-30B-A3B | 0.894 | 0.879 | [0.868, 0.901] | 0.834 |
| LLM-jp-4-32B | 0.889 | 0.877 | [0.867, 0.898] | 0.832 |
| LLM-jp-3.1-13B | 0.899 | 0.863 | [0.857, 0.901] | 0.833 |
| LLM-jp-3.1-1.8B | 0.863 | 0.804 | [0.807, 0.856] | 0.775 |

*Precision and recall use relaxed matching; Macro48 is also based on relaxed matching. Confidence intervals use 2,000 document-level bootstrap resamples of the 36 test documents. LLM limits match Table 2: 4,096 tokens except 2,600 for the two LLM-jp-3.1 models. These intervals describe seed 42, not the uncertainty of the three-seed mean.*

**Supplementary Table S3. Zero-shot output diagnostics (prompt-only ZS and constrained ZS+C, 3,072-token limit).**

| **Model** | **Output / context limit** | **ZS: unparseable** | **ZS: out-of-inventory labels** | **ZS: truncated** | **ZS: predictions per sentence** | **ZS: P / R** | **ZS+C: truncated** | **ZS+C: P / R** |
|---|---|---|---|---|---|---|---|---|
| MedGemma-27B | 3072 / 8192 | 0 | 218 | 2 | 17.9 | 0.448 / 0.269 | 6 | 0.433 / 0.258 |
| Gemma-3-27B | 3072 / 8192 | 0 | 235 | 0 | 15.1 | 0.475 / 0.240 | 0 | 0.456 / 0.232 |
| Llama-3.3-70B | 3072 / 8192 | 0 | 173 | 0 | 10.8 | 0.515 / 0.186 | 0 | 0.508 / 0.184 |
| Qwen3-30B-A3B | 3072 / 8192 | 3 | 236 | 24 | 18.0 | 0.312 / 0.188 | 25 | 0.319 / 0.196 |
| LLM-jp-4-32B | 3072 / 8192 | 17 | 95 | 35 | 4.1 | 0.386 / 0.093 | 0 | 0.000 / 0.000 |
| Med42-70B | 3072 / 8192 | 7 | 569 | 0 | 8.8 | 0.296 / 0.088 | 0 | 0.338 / 0.090 |
| LLM-jp-3.1-13B | 3072 / 4096 | 2 | 233 | 152 | 22.1 | 0.045 / 0.033 | 90 | 0.057 / 0.026 |
| LLM-jp-3.1-1.8B | 3072 / 4096 | 11 | 1486 | 210 | 61.1 | 0.012 / 0.025 | 159 | 0.011 / 0.018 |
| SIP-jmed-13B | 3072 / 8192 | 92 | 47 | 147 | 1.7 | 0.122 / 0.014 | 99 | 0.039 / 0.010 |

*Reference sentences contain 29.8 entities on average. Unparseable: no JSON array could be recovered. Out-of-inventory: predictions whose label is not among the 109 prompt labels (scored as false positives). Under ZS+C neither can occur by construction.*

Supplementary Table S4. Recall and output-length diagnostics for the 900-token condition (seed 42).

| Model | Truncated sentences | Share of reference entities | Entities per sentence (truncated / other) | Recall, truncated | Recall, other | Precision (truncated / other) | Share of false negatives in truncated sentences |
|---|---|---|---|---|---|---|---|
| Llama-3.3-70B | 86 | 52% | 79 / 18 | 0.581 | 0.932 | 0.909 / 0.930 | 87% |
| Med42-70B | 87 | 53% | 78 / 18 | 0.584 | 0.935 | 0.908 / 0.937 | 88% |
| MedGemma-27B | 81 | 50% | 80 / 18 | 0.578 | 0.927 | 0.901 / 0.931 | 86% |
| Gemma-3-27B | 84 | 52% | 80 / 18 | 0.588 | 0.934 | 0.905 / 0.936 | 87% |
| SIP-jmed-13B | 84 | 52% | 80 / 18 | 0.580 | 0.923 | 0.904 / 0.929 | 85% |
| Qwen3-30B-A3B | 85 | 52% | 79 / 18 | 0.572 | 0.915 | 0.893 / 0.926 | 84% |
| LLM-jp-4-32B | 79 | 50% | 81 / 18 | 0.572 | 0.910 | 0.886 / 0.917 | 82% |
| LLM-jp-3.1-13B | 82 | 51% | 80 / 18 | 0.573 | 0.917 | 0.897 / 0.926 | 84% |
| LLM-jp-3.1-1.8B | 81 | 50% | 80 / 18 | 0.526 | 0.873 | 0.853 / 0.899 | 79% |
| UTH-BERT (same split as Med42) | 87 | 53% | 78 / 18 | 0.798 | 0.848 | 0.812 / 0.878 | — |

*The UTH-BERT comparison uses the same test-sentence split as Med42-70B. These are descriptive diagnostics; sentences were classified by the LLM output reaching its token limit.*

**Supplementary Table S5. Five-point generation-budget analysis (seed 42).**

| Model | 512 | 900 | 1,024 | 2,048 | Upper limit |
|---|---|---|---|---|---|
| Llama-3.3-70B | 0.711 (173) | 0.826 (86) | 0.846 (65) | 0.894 (11) | 0.911 (2) |
| Med42-70B | 0.712 (173) | 0.828 (87) | 0.846 (67) | 0.893 (11) | 0.908 (2) |
| MedGemma-27B | 0.713 (171) | 0.826 (81) | 0.844 (64) | 0.886 (11) | 0.902 (2) |
| Gemma-3-27B | 0.716 (170) | 0.831 (84) | 0.850 (66) | 0.897 (11) | 0.912 (2) |
| SIP-jmed-13B | 0.708 (171) | 0.823 (84) | 0.841 (66) | 0.883 (11) | 0.896 (3) |
| Qwen3-30B-A3B | 0.698 (171) | 0.815 (85) | 0.831 (68) | 0.872 (10) | 0.887 (3) |
| LLM-jp-4-32B | 0.708 (169) | 0.815 (79) | 0.834 (62) | 0.871 (9) | 0.883 (2) |
| LLM-jp-3.1-13B | 0.706 (170) | 0.819 (82) | 0.836 (62) | 0.876 (11) | 0.881 (6) |

| Model | 512 | 900 | 1,024 | 2,048 | Upper limit |
|---|---|---|---|---|---|
| LLM-jp-3.1-1.8B | 0.676 (165) | 0.780 (81) | 0.796 (65) | 0.830 (13) | 0.832 (10) |

*Each entry gives relaxed F1 (number of truncated sentences). Column headings are output-token limits. The upper limit is 4,096 tokens for seven models and 2,600 for LLM-jp-3.1-13B and LLM-jp-3.1-1.8B. All scores use the same seed-42 checkpoint within each model.*

Supplementary Table S6. Seed-specific paired relaxed-F1 differences between medical and general-purpose LLMs after fine-tuning at matched larger budgets.

| Medical–general pair | Seed | Δ relaxed F1 | 95% paired document-bootstrap CI |
|---|---|---|---|
| 13B | 42 | +0.007 | [+0.000, +0.014] |
| 13B | 1 | +0.007 | [+0.002, +0.012] |
| 13B | 2 | +0.008 | [+0.001, +0.015] |
| 27B | 42 | −0.010 | [−0.016, −0.004] |
| 27B | 1 | +0.006 | [+0.001, +0.011] |
| 27B | 2 | +0.007 | [−0.002, +0.020] |
| 70B | 42 | −0.004 | [−0.010, +0.003] |
| 70B | 1 | −0.004 | [−0.010, +0.003] |
| 70B | 2 | −0.004 | [−0.009, +0.000] |

*Positive values favor the medical model. The 13B comparison uses 2,600 tokens for both models; the 27B and 70B comparisons use 4,096. Rounded endpoints shown as 0.000 do not imply equivalence.*

Supplementary Table S7. Label-level relaxed and exact-surface F1 for Med42-70B and UTH-BERT (seed 42).

| Label | Test entities | Med42-70B relaxed | Med42-70B exact-surface | UTH-BERT relaxed | UTH-BERT exact-surface |
|---|---|---|---|---|---|
| state-patient | 2318 | 0.91 | 0.85 | 0.79 | 0.71 |
| clinical_test | 587 | 0.93 | 0.88 | 0.81 | 0.74 |
| treatment | 614 | 0.91 | 0.84 | 0.85 | 0.79 |
| drug | 283 | 0.95 | 0.94 | 0.81 | 0.77 |
| PN-Negative-patient | 276 | 0.95 | 0.92 | 0.94 | 0.90 |
| state-others | 155 | 0.56 | 0.53 | 0.45 | 0.40 |
| device | 75 | 0.81 | 0.65 | 0.58 | 0.52 |
| time_span | 59 | 0.67 | 0.65 | 0.64 | 0.58 |
| value-others | 63 | 0.28 | 0.28 | 0.29 | 0.26 |

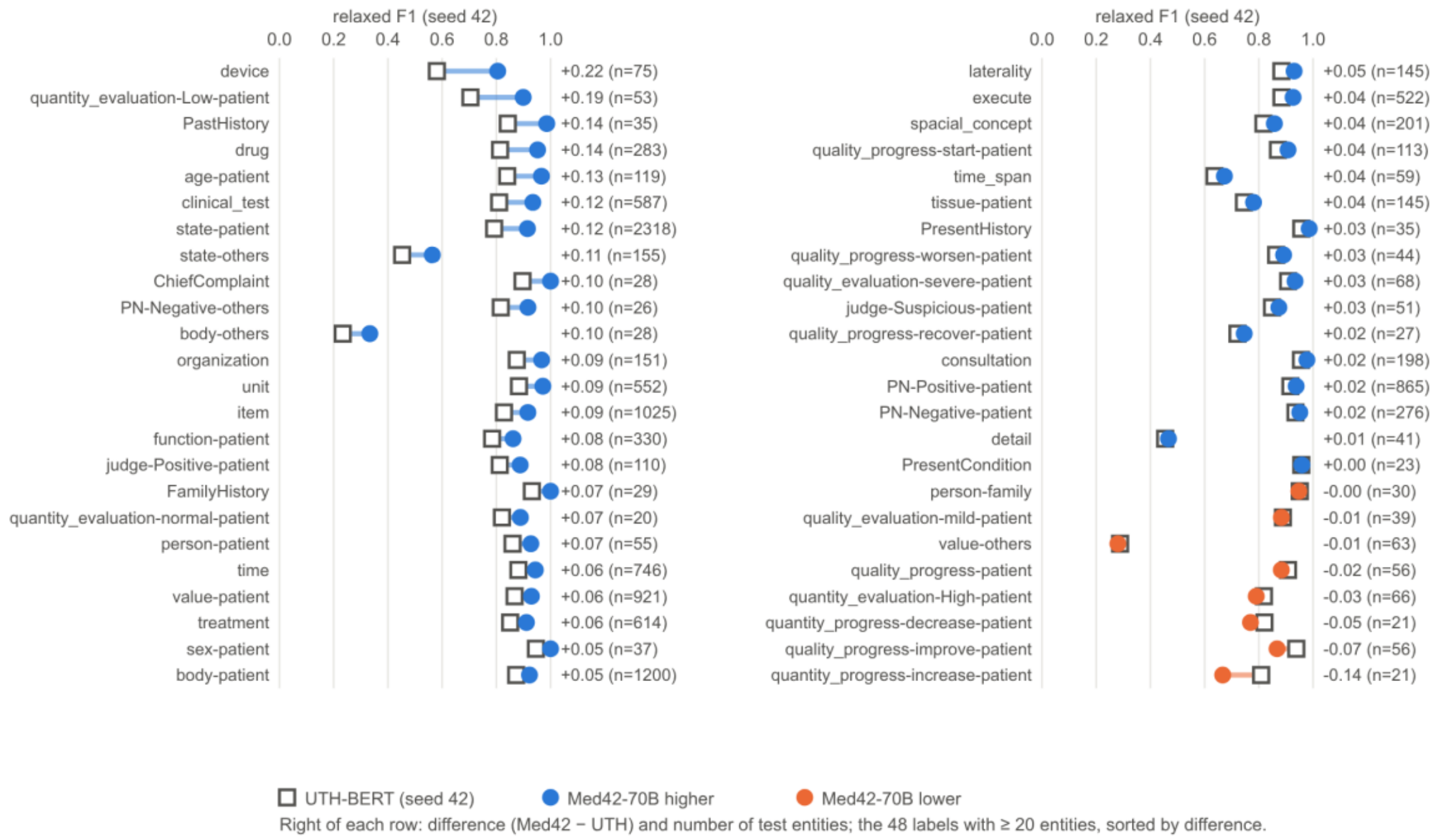

Supplementary Figure S1. Label-level relaxed F1 for Med42-70B versus UTH-BERT across 48 test labels with at least 20 reference entities (seed 42). Filled circles show Med42-70B, squares show UTH-BERT; blue and orange indicate higher and lower Med42-70B F1, respectively. The differences (Med42 minus UTH) and reference counts are shown beside each row.

**Supplementary Table S8. Selected label-level recall and relaxed F1 for Med42-70B (seed 42; larger output limit).**

| Entity label | Reference pairs (n) | Recall | Relaxed F1 |
|---|---|---|---|
| PN-Negative-patient | 276 | 0.949 | 0.951 |
| PN-Negative-others | 26 | 0.846 | 0.917 |
| state-patient | 2318 | 0.924 | 0.915 |
| state-others | 155 | 0.413 | 0.564 |
| value-patient | 921 | 0.943 | 0.929 |
| value-others | 63 | 0.222 | 0.280 |

These values use the same seed-42 4,096-token predictions and matching rules as Supplementary Table S7. The full comparison of 48 frequent labels is shown in Supplementary Figure S1.

**Supplementary Table S9. Representative model-level annotation and entity-boundary errors from Med42-70B seed-42, 4,096-token predictions.**

| Test ID | Text excerpt | Reference annotation | Model prediction | Diagnostic interpretation |
|---|---|---|---|---|
| 010_4_4 | "NGSP 値 3" | value-others: "3" | value-patient: "3" | Label assignment differs from the reference; the annotation convention requires examination. |
| 039_3_7 | "真皮浅層" | tissue-patient: "真皮浅層" | tissue-patient: "真皮"; spacial_concept: "浅層" | The reference tissue span is fragmented into two predictions, with the second assigned a different label. |

The examples illustrate disagreements with the reference annotations and do not constitute an exhaustive error taxonomy or a frequency analysis.